\pdfoutput=1
\documentclass[11pt]{article}

\usepackage[final]{acl}

\usepackage{times}
\usepackage{latexsym}

\usepackage[T1]{fontenc}
\usepackage[utf8]{inputenc}

\usepackage{microtype}
\usepackage{inconsolata}
\usepackage{booktabs}
\usepackage{soul}
\usepackage{color}

\usepackage{multirow}
\usepackage{multicol}
\usepackage{enumitem,kantlipsum}
\usepackage[normalem]{ulem}
\usepackage{color,colortbl}
\usepackage{todonotes}
\usepackage{soul}
\usepackage{threeparttable, makecell}
\usepackage{makecell}
\usepackage{xcolor}

\usepackage{tcolorbox}

\usepackage{comment}
\usepackage{array}
\usepackage{lscape}
\usepackage{multirow}
\usepackage{graphicx}
\usepackage[normalem]{ulem}
\useunder{\uline}{\ul}{}

\usepackage{array}
\usepackage{multirow}
\usepackage{geometry}
\usepackage{algorithm}
\usepackage{algpseudocode}
\usepackage{amsmath}
\usepackage{subcaption}
\usepackage{fancybox}
\usepackage{tikz}
\usepackage{listings}
\usepackage{color}

\usepackage{pifont}% http://ctan.org/pkg/pifont

\usepackage{enumitem}
\setlist{nosep,topsep=-\parskip}

\usepackage{soul}
\usepackage{xcolor}
\definecolor{dkgreen}{rgb}{0,0.6,0}

\usepackage{listings}
\usepackage{graphicx}
\definecolor{codegray}{gray}{0.9}

\lstdefinestyle{pythonstyle}{
    backgroundcolor=\color{codegray},   
    language=Python,
    basicstyle=\ttfamily\footnotesize,
    keywordstyle=\color{blue},
    stringstyle=\color{red},    
    breaklines=true,
    frame=single,
    keepspaces=true,
    showstringspaces=false,
}

\usepackage{url}

\usepackage{graphicx}
\definecolor{lightblue}{rgb}{.50,.95,1}
\definecolor{tri}{rgb}{.25,.88,.82}
\definecolor{lilac}{rgb}{0.85,0.64,0.85}

\newcommand{\cultranai}{\textbf{\emph{CultranAI}}}

\usepackage{amssymb}
\usepackage{hyperref}
\usepackage{verbatim}
\usepackage{setspace}

\usepackage{colortbl}
\usepackage{booktabs}
\usepackage{graphicx}
\usepackage{mathrsfs}
\usepackage{xcolor}
\definecolor{mycolor}{RGB}{0, 110, 180}

\usepackage{arabtex}
\usepackage{utf8}
\setcode{utf8}

\title{\textcolor{mycolor}{\textsf{\textit{CultranAI}}} at PalmX 2025: Data Augmentation for Cultural Knowledge Representation
}

\author{
  Hunzalah Hassan Bhatti$^{1}$\thanks{~The contribution was made while the author was interning at the Qatar Computing Research Institute.}, Youssef Ahmed$^{1}$$^{*}$, Md Arid Hasan$^2$, Firoj Alam$^{3}$ \\
  $^{1}$Qatar University, $^{2}$University of Toronto, Canada \\
  $^{3}$Qatar Computing Research Institute \\  
  \texttt{hunzalahhassan@gmail.com, fialam@hbku.edu.qa}
}

\begin{document}
\maketitle

\begin{abstract}
In this paper, we report our participation to the PalmX cultural evaluation shared task. Our system, \cultranai{}, focused on data augmentation and LoRA fine-tuning of large language models (LLMs) for Arabic cultural knowledge representation. We benchmarked several LLMs to identify the best-performing model for the task. In addition to utilizing the PalmX dataset, we augmented it by incorporating the Palm dataset and curated a new dataset of over 22K culturally grounded multiple-choice questions (MCQs). Our experiments showed that the Fanar-1-9B-Instruct model achieved the highest performance. We fine-tuned this model on the combined augmented dataset of 22K+ MCQs. On the blind test set, our submitted system ranked 5th with an accuracy of 70.50\%, while on the PalmX development set, it achieved an accuracy of 84.1\%. We made experimental scripts publicly available for the community.\footnote{\url{https://github.com/hunzed/CultranAI}}
\end{list}
% This new dataset was generated based on queries derived from PalmX, reflecting the cultural context more comprehensively. 
\end{abstract}

\section{Introduction}
\label{sec:introduction}

Cultural information plays a pivotal role in shaping human identity, behavior, and social interactions. It encompasses the shared beliefs, values, customs, languages, traditions, and collective knowledge of a community or society. In today’s interconnected information, communication, and interaction ecosystem, hundreds of millions of users engage with LLMs for everyday queries - many of which involve aspects of local culture, traditions, cuisine, and more~\cite{pawar2024survey,hasan-etal-2025-nativqa}. A central challenge lies in evaluating how effectively LLMs comprehend and generate responses to such culturally embedded queries, particularly in multilingual settings characterized by significant dialectal variation. Other challenges include how to develop culturally aligned LLMs~\cite{wang2023aligning} and make them available in low-compute environments~\cite{hu2022lora}. Recent initiatives have introduced evaluation resources - such as culturally relevant datasets, task-specific benchmarks, and performance metrics - to assess LLM capabilities in this domain \cite{myung2024blend,NEURIPS2024_77f089cd,mousi-etal-2025-aradice}.
% these efforts remain limited in scope~\cite{myung2024blend,NEURIPS2024_77f089cd,mousi-etal-2025-aradice}.

Yet these efforts remain limited, especially in achieving deeper, dialect-specific advancements. Addressing this gap requires sustained, targeted, rigorous initiatives. The PalmX Shared Task at ArabicNLP 2025~\cite{alwajih2025palmx} is a step in this direction, offering a dedicated benchmark for culturally specific evaluation with a special emphasis on Arabic - thereby advancing the development of LLMs that are both linguistically and culturally aligned. Other recent relevant efforts for Arabic include the development of Arabic-centric LLMs~\cite{fanar2024,sengupta2023jais,bari2024allam}, leaderboards~\cite{almatham2025balsamplatform},
% \footnote{e.g., \url{https://benchmarks.ksaa.gov.sa/b/balsam}} 
and culturally specific datasets~\cite{alwajih-etal-2025-palm,ayash2025saudiculture}. 

To advance the state of the art in Arabic cultural knowledge representation within LLMs, in this paper, we report our participation in the shared task. We specifically focus on the cultural evaluation subtask. To address the challenges of training and deploying LLMs in low-compute resource settings, we conducted a comparative analysis of quantized vs. full-precision models. In parallel, we employed LLM-driven data augmentation strategies to improve the model accuracy.
To summarise, the contributions of our study are as follows. 
\begin{itemize}[noitemsep,topsep=0pt,leftmargin=*,labelsep=.5em] 
 \item We provide a performance comparison of different LLMs (Arabic-centric and multilingual) in a zero-shot setup.
 \item We demonstrate that the performance gap between quantized models and their full-precision counterparts is minimal.
 \item We show that data augmentation contributes to improving model performance. 
\end{itemize}

% \textit{\textbf{Our findings}}

\section{Related Work}
\label{sec:related work}

\paragraph{General Capabilities of LLMs.}

LLMs have demonstrated remarkable capabilities across a wide range of natural language processing (NLP) tasks, including text classification, question answering, summarization, and dialogue generation~\cite{bubeck2023sparks,abdelali-etal-2024-larabench}. Their ability to leverage vast amounts of pretraining data and adapt to downstream tasks with minimal supervision has enabled strong performance in both zero-shot and few-shot settings~\cite{abdelali-etal-2024-larabench}. These advances have accelerated the integration of LLMs into diverse real-world applications spanning education, healthcare, finance, and customer support.

\paragraph{Cultural and Everyday Knowledge.}
Despite successes in several downstream NLP tasks, LLMs often underperform on tasks requiring culturally grounded knowledge, particularly in low-resource languages and dialects~\cite{pawar2024survey,hasan-etal-2025-nativqa,alam2025nativqaframeworkenablingllms}. A culturally aligned model should accurately interpret and generate content that reflects local linguistic forms, social norms, and lived experiences across domains such as healthcare, education, and cuisine~\cite{NEURIPS2024_77f089cd,li2024culturellm,shi-etal-2024-culturebank}. However, current models frequently fail to capture region-specific expressions and indigenous knowledge, limiting their effectiveness in culturally nuanced contexts~\cite{myung2024blend,chiu2024culturalbench}. To address these limitations, recent research has focused on developing benchmarks and datasets that evaluate and enhance LLMs’ performance for both cultural and everyday information-seeking queries. These resources span mono- and multilingual settings and are sourced from diverse origins, including Wikipedia~\cite{yang-etal-2018-hotpotqa,kwiatkowski-etal-2019-natural}, Google Search QA~\cite{khashabi-etal-2021-gooaq-open}, Reddit forums~\cite{fan-etal-2019-eli5}, and native speaker-authored question–answer pairs~\cite{clark-etal-2020-tydi}. Other approaches combine native and machine-translated content or employ LLMs to generate culturally relevant QA datasets~\cite{putri2024can,NEURIPS2024_77f089cd}.

Although English and multilingual resources have advanced the state of the art in culturally aligned LLMs, the richness and diversity of the Arabic language and its dialects require dedicated efforts in both resource creation and culturally aligned model development. Recent initiatives have begun addressing this gap through the development of datasets for benchmarking and fine-tuning Arabic-centric models~\cite{mousi-etal-2025-aradice,alwajih-etal-2025-palm}. The PalmX Shared Task at ArabicNLP 2025 is a targeted initiative to advance culturally aligned LLM development through a benchmark for culturally grounded evaluation in Arabic.  

\section{Task and Dataset}
\label{sec:dataset}

\subsection{Task Overview}
The PalmX 2025 shared task offered two subtaks, one of which is \textit{General Culture Evaluation} (Subtask 1). The goal of the task is to benchmark Arabic language models on their ability to answer culturally grounded multiple-choice questions in Modern Standard Arabic (MSA). The questions span various domains such as history, customs, geography, literature, and food, and are designed to reflect general cultural literacy in Arab countries.

Participants are provided with a training and development set of MCQs, each with four answer options. The final evaluation is performed on a held-out test set of 2,000 questions, with accuracy as the primary metric. The task encouraged the use of external data for model enhancement, provided that models remain under 13 billion parameters and final checkpoints are submitted for evaluation.
% This setup aimed to assess both the breadth and depth of cultural knowledge in Arabic LLMs, while encouraging diverse strategies for model enhancement within the specified resource constraints. 
% Evaluation was based on accuracy, measured as the percentage of correctly answered questions on the held-out test set.

% We have participated in \textit{subtask 1: culture evaluation}, Our contribution goes beyond baseline fine-tuning by introducing a novel data augmentation strategy: we leverage the NativQA framework~\cite{alam2025nativqaframeworkenablingllms} to generate culturally relevant QA pairs, convert them into multiple-choice format, and integrate them into training. This approach expands the scope and regional depth of the original dataset, offering a stronger foundation for culturally informed model performance.

\subsection{Dataset}

% Our work is based primarily on PalmX, the data set released for the PalmX2025 shared task (Subtask 1) \cite{palmxcompetition}, which focuses on evaluating cultural knowledge in Arabic through multiple-choice question answering. 
% \paragraph{PalmX.} 
The \textit{PalmX 2025 Cultural Evaluation} dataset consists of 2,000 training examples and 500 development examples, each formulated as a MCQ with four answer options and a single correct answer. 
% All content is written in MSA, and the questions are designed to assess general cultural literacy in the Arab world. 
We used the training split for fine-tuning and reserved the development set for evaluation, except for our final iterations, where we use both training and evaluation splits for fine-tuning. 
% The data set is publicly accessible via Hugging Face after registration.

\subsection{Data Augmentation}

\paragraph{Palm.} To complement PalmX dataset, we incorporated the Palm dataset \cite{alwajih-etal-2025-palm}, a broader community-curated resource created by contributors from the 22 Arab countries. Unlike PalmX, which is entirely in MSA, Palm spans both MSA and various dialects, offering instruction-style QA pairs on 20 culturally relevant topics such as heritage, cuisine, history, and proverbs. All examples are manually written by native speakers with cultural familiarity, ensuring authenticity and regional diversity. Although Palm includes training and test splits, only the test portion, comprising 1,926 QA pairs, is publicly available.

We split the available Palm test set into two halves: one for fine-tuning, and the other for evaluation. The splits were created using stratified sampling based on country, ensuring balanced representation across regions in both halves. To bring its free-form QA format in line with PalmX, we converted each example into MCQ format using GPT-4.1. Specifically, we generated three plausible distractors per question, preserving semantic coherence and cultural plausibility. 
% This transformation allowed us to reuse high-quality instructional content while maintaining consistency in format and difficulty across datasets. 
In Appendix~\ref{lst:prompt_for_country}, we provided the prompt that we used for MCQ version of the palm dataset. 

% \subsection{Data Augmentation via NativQA Framework}

\paragraph{Extending PalmX Dataset.} 
% \todo[inline]{FA: working}

To further diversify and expand our training data, we leveraged the NativQA framework~\cite{alam2025nativqaframeworkenablingllms} in combination with GPT-4.1. The NativQA framework can seamlessly curate large-scale QA pairs based on user queries, ensuring cultural and regional alignment in native languages. GPT-4.1 was selected for its optimal trade-off between cost and performance at the time of experimentation. In all cases, we employed zero-shot prompting with GPT-4.1.

\begin{figure}
\centering
\includegraphics[width=1.0\linewidth]{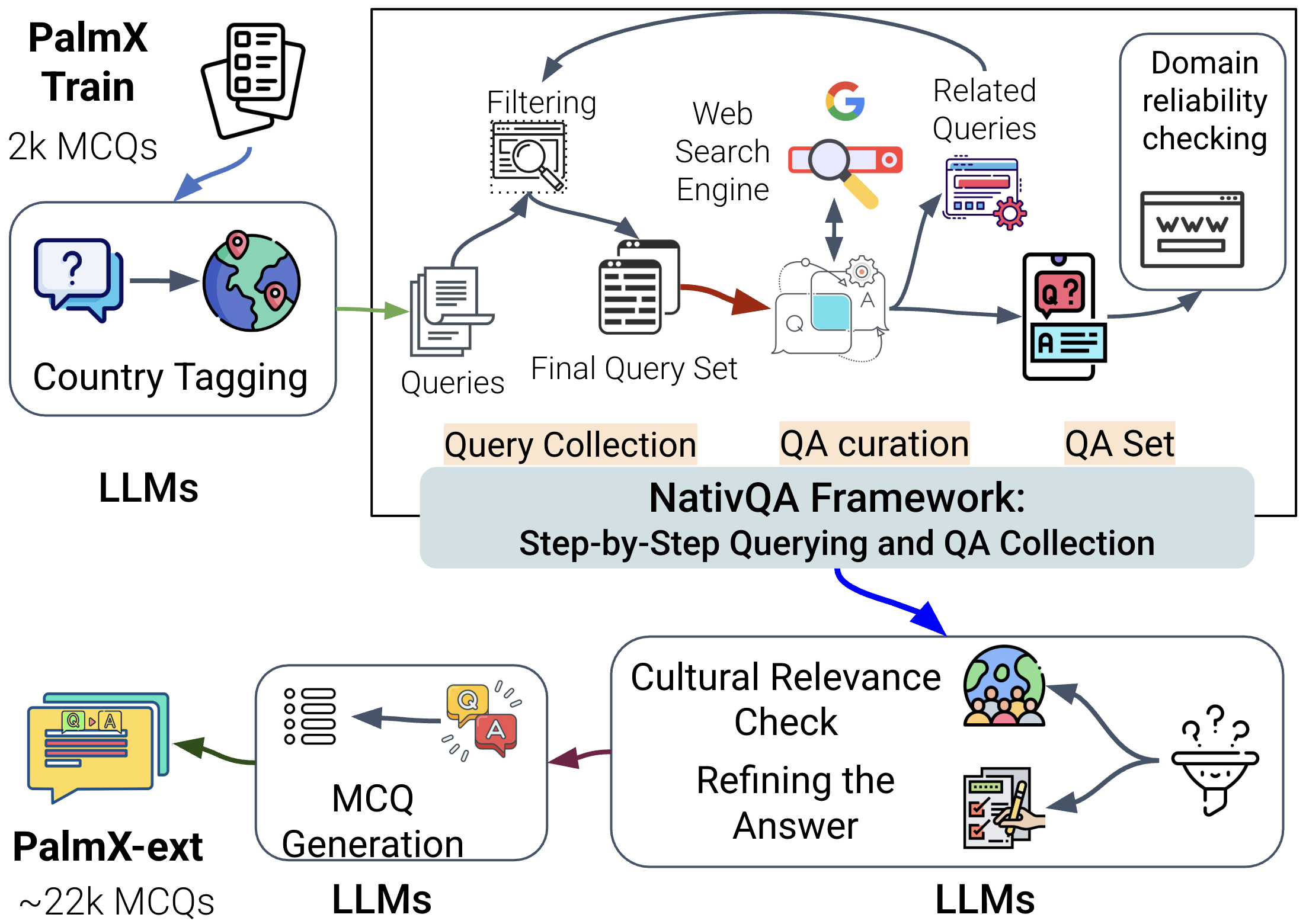}
\caption{Pipeline for extending the PalmX dataset using the NativQA framework and GPT-4.1.}
% \vspace{-0.2cm}
\label{fig:palmx_ext_pipeline}
\vspace{-0.2cm}
\end{figure}

As illustrated in Figure~\ref{fig:palmx_ext_pipeline}, our process for extending the PalmX dataset began by identifying the country associated with each question using GPT-4.1. The prompt used for this task is provided in Listing~\ref{lst:prompt_for_country}. This country information was then combined with the NativQA framework to curate location-specific QA pairs.  

The NativQA framework’s retrieval process was carried out in two iterations to maximize topical diversity. To maintain factual quality, all answers were filtered using NativQA’s Domain Reliability Check (DRC), which retains only those sourced from NativQA-verified web domains. Furthermore, GPT-4.1 was employed to filter and refine the answers described in Listing~\ref{lst:prompt_for_filtering_palm}. The idea is to remove culturally irrelevant or factually incorrect QA pairs and refine answers for conciseness and the overall quality of the dataset. Similar to the Palm test set, these new entries were converted into MCQ format to match PalmX, using the same prompt applied to the original Palm data. This process augmented the original dataset with culturally rich examples while preserving structural and contextual consistency with the PalmX questions. We also manually reviewed 50 samples, which received an average score of about 7.4 on a scale of 10 for accuracy and clarity. We refer to this dataset as the PalmX-ext set. In Table \ref{tab:data_distribution}, we report the distribution of the dataset that we used for training and evaluation.

% \todo[inline]{Need a better figure. Please check if you can draw on google doc. Here is a raw copy from nativqa: \url{https://docs.google.com/presentation/d/1fwEBgxdVhOFhFjpaP0QJiS69oaQc59HR/edit?usp=sharing&ouid=106855749914204196680&rtpof=true&sd=true}}

\begin{table}[]
\centering
\setlength{\tabcolsep}{3pt}
\scalebox{0.8}{%
\begin{tabular}{lrrr}
\toprule
\multicolumn{1}{c}{\textbf{Data}} & \multicolumn{1}{c}{\textbf{Train}} & \multicolumn{1}{c}{\textbf{Dev}} & \multicolumn{1}{c}{\textbf{Test}} \\ \midrule
PalmX & 2,000 & 500 & 2,000 \\
Palm & 950 & 950 & - \\
PalmX-ext & 22,000 & - &  -\\ \bottomrule
\end{tabular}
}
\vspace{-0.2cm}
\caption{Distribution of the datasets used for training, development and test.}
\label{tab:data_distribution}
\vspace{-0.2cm}
\end{table}

\section{Experiments}
% need to mention all the models used and cite

\noindent
\textbf{Models.}
We began by evaluating a set of open-sourced LLMs in a zero-shot setup on both evaluation datasets. This initial comparison helped us identify the most promising model for fine-tuning, and further demonstrated the utility of the Palm test set as an effective evaluation benchmark.

To identify the most suitable model for the task, we evaluated a set of models based on their performance on the PalmX development set. The models for experiments include \texttt{tiny-random-LlamaForCausalLM},\footnote{\url{https://huggingface.co/HuggingFaceH4/tiny-random-LlamaForCausalLM}} \texttt{Qwen2.5-7B-Instruct}~\cite{wang2024qwen2}, \texttt{Jais-13B-Chat}~\cite{sengupta2023jais}, \texttt{Miraj Mini},\footnote{\url{https://huggingface.co/arcee-ai/Meraj-Mini}} \texttt{Llama-3.1-8B-Instruct}~\cite{touvron2023llama}, \texttt{NileChat-3B}~\cite{mekki2025nilechat}, \texttt{ALLaM-7B-Instruct}~\cite{bari2024allam}, and \texttt{Fanar-7B-Instruct}~\cite{fanar2024}. We selected both Arabic-centric and multilingual models to compare the effectiveness of models tailored to Arabic with those trained on broader multilingual corpora. The \texttt{tiny-random-LlamaForCausalLM} model was included for baseline results. 
\noindent
\textbf{Training Setup.}
We experimented with two fine-tuning approaches: LoRA and QLoRA. LoRA trained only a set of low-rank adapter layers while keeping the rest of the model frozen, whereas QLoRA combined 4-bit quantization with LoRA adapters to reduce memory usage without a substantial drop in performance. 
% For LoRA, we set the rank $r = 64$, $\alpha = 16$, and a dropout of $0.05$.

Both methods were trained on the same mix of datasets: PalmX Train, Palm, and PalmX-ext. We used Fanar’s native tokenizer with its default tokenization strategy, a batch size of $4$, and gradient accumulation steps of $4$. Training was conducted for $3$ epochs with a learning rate of $2\times 10^{-4}$, saving the best-performing checkpoint at the end of each run. Fine-tuning followed a specific prompt %format for the Arabic MCQ task-
- each question was prefixed by a system prompt, followed by the question and answer choices, as shown in Figure~\ref{fig:formatted_prompt_structure}.
% (see Figure~\ref{fig:formatted_prompt_structure} for an example of a formatted prompt).

\noindent
\textbf{Data Augmentation.}
We then studied the effect of our data augmentation strategy by comparing LoRA training on \textbf{PalmX Train} alone \textit{vs.} LoRA training on \textbf{PalmX Train} combined with our augmented \textbf{Palm} and \textbf{PalmX-ext} datasets. This experiment used the same configuration as the earlier LoRA \textit{vs.} QLoRA comparison. After identifying the best-performing approach, we performed hyperparameter tuning to optimise its performance. In Appendix~\ref{appendix:tuning_experiments} and \ref{appendix:tuning_results}, we report complete experimental setup and results, respectively. 
% The complete experimental setup is detailed in Appendix~\ref{appendix:tuning_experiments}, and the results are reported in Appendix~\ref{appendix:tuning_results}.

% \todo[inline]{discuss about the model that has been submitted}
\noindent
% \textbf{Submitted model.}
After identifying the best-performing model, the most effective fine-tuning strategy, and the optimal hyperparameters, the final submission was trained for 3 epochs with a learning rate of $2\times10^{-4}$, LoRA rank $64$, dropout $0.1$, and scaling factor $\alpha=16$, using PalmX Train and Dev, Palm, and PalmX-ext.  

\section{Results}
\label{sec:results}

\noindent
\textbf{Zero-shot Performance.}  
Table~\ref{tab:base-models} reports the zero-shot performance of several multilingual and Arabic-centric instruction models. \textbf{Fanar-7B} achieved the highest accuracy on PalmX Dev, making it our choice for fine-tuning.

\begin{table}[!tbh]
\centering
\small
\setlength{\tabcolsep}{3pt}
\scalebox{0.92}{%
\begin{tabular}{lcc}
\toprule
\textbf{Model} & \textbf{PalmX Dev} & \textbf{Palm} \\
\midrule
tiny-random-Llama    & 23.40 & 26.51 \\
Qwen2.5-7B-Inst.      & 69.20 & 74.32 \\
Jais-13B-chat         & 61.00 & 55.72 \\
Miraj Mini            & 70.20 & \textbf{75.99} \\
Llama3.1-8B-Inst.     & 66.60 & 74.06 \\
Nilechat-3B           & 70.00 & 66.89 \\
ALLaM-7B-Inst.        & 70.60 & 74.32 \\
\textbf{Fanar-7B}     & \textbf{72.40} & 73.34 \\
\bottomrule
\end{tabular}
}
\vspace{-0.2cm}
\caption{Zero-shot performance of base models.
% on PalmX Dev and Palm Eval Split.
}
\label{tab:base-models}
\vspace{-0.2cm}
\end{table}

\noindent
\textbf{Comparison on PEFT methods.}  
Table~\ref{tab:finetune-techniques} compares LoRA with its quantized variant (QLoRA) under identical settings. LoRA achieved a slight improvement over QLoRA on PalmX Dev, suggesting that full-precision adapters were marginally more effective.

\begin{table}[!tbh]
  \centering
\setlength{\tabcolsep}{3pt}
\scalebox{0.85}{%
  \begin{tabular}{lc}
    \toprule
    \textbf{Method} & \textbf{PalmX Dev (\%)} \\
    \midrule
    QLoRA (4-bit)               & 80.00 \\
    LoRA                        & \textbf{80.60} \\
    \bottomrule
  \end{tabular}
}
\vspace{-0.2cm}
\caption{Results using PEFT methods.}
\label{tab:finetune-techniques}
\vspace{-0.2cm}
\end{table}

\noindent
\textbf{Effect of Data Augmentation.}  
Table~\ref{tab:dataset_comparison} evaluates the impact of adding augmented Palm and PalmX-ext data to PalmX Train. The augmented dataset led to substantial gains on PalmX Dev, indicating improved generalization. A more detailed error analysis is provided in Appendix~\ref{sec:aug-analysis}.

\begin{table}[h]
\centering
\small
\setlength{\tabcolsep}{3pt}
\scalebox{0.92}{%
\begin{tabular}{lcc}
\toprule
\textbf{Training Data} & \textbf{PalmX Dev (\%)} \\
\midrule
PalmX             & 76.6 \\
PalmX + PalmX-ext + Palm & \textbf{80.6}  \\
\bottomrule
\end{tabular}
}
\vspace{-0.2cm}
\caption{Results with and without augmented data.}
\label{tab:dataset_comparison}
\vspace{-0.2cm}
\end{table}

The final submitted model achieved an accuracy of 84.1\% on the Palm test set. As the PalmX development set was included in the training data, it was excluded from evaluation on the submitted model. On the blind test set, the model obtained an accuracy of 70.5\%. A more detailed analysis of the discrepancy between Dev and Test performance is provided in Appendix~\ref{sec:dev-test-analysis}.

\section{Conclusions and Future Work}
\label{sec:conclusions}
In this paper, we present our system, \textit{CultranAI}, designed to enhance cultural knowledge representation in LLMs for Arabic. We conduct an extensive comparative evaluation in a zero-shot setting using various multilingual and Arabic-centric models, which led us to identify \textit{Fanar} as the most suitable model for further experimentation. To assess performance in low-compute scenarios, we explored different PEFT methods. We also investigated data augmentation techniques aimed at improving model accuracy. Our proposed system achieved an accuracy of 84.1\%  on the Palm set, and ranked $5^{th}$ on the blind test set with an accuracy of 70.5\%. Future work will focus on refining data augmentation pipelines and further exploring model generalizability.

% and exploring more organic ways of gathering data that preserve cultural context. Our findings suggest that combining culturally aligned augmentation with targeted iterative training is a promising path for advancing LLM performance in specialized cultural domains.

% We presented CultranAI, our system for the PalmX 2025 Shared Task on Arabic cultural knowledge representation, combining extensive dataset augmentation with LoRA fine-tuning of the Fanar-7B model. By integrating Palm, PalmX, and a newly curated PalmX-ext dataset of culturally grounded MCQs, our best configuration achieved 84.1\% on the Palm set, and ranked 5th in the blind evaluation with 70.5\% accuracy on the PalmX test set. Expanding the dataset more than tenfold brought clear improvements, showing the strong impact of well-targeted augmentation on model performance.
% Future directions include refining augmentation pipelines and exploring more organic ways of gathering data that preserve cultural context. Our findings suggest that combining culturally aligned augmentation with targeted iterative training is a promising path for advancing LLM performance in specialized cultural domains.

\section{Limitations}
\label{sec:limitations}
While augmentation brought clear improvements, we believe the performance could have been higher with more careful dataset preparation. In the Palm dataset, instructional QAs were directly converted to MCQ, but some QAs exceeded the 512-token PalmX limit. PalmX-ext avoided this by reformatting MCQs in the first post-processing step. Another problem was distractor quality: in both PalmX-ext and Palm, distractors were often shorter than the correct answer. These issues can be addressed by refining the prompts for distractor generation and adding a processing step to truncate long Palm QAs. 
% However, reprocessing the dataset was not feasible in our case due to high costs.

\bibliography{bib/bibliography}

\appendix

% \section{Example Appendix}
% \label{sec:appendix}

% \section{Examples}

\newpage
\section{Prompts}

\begin{figure}[h]
\centering
\begin{quote}
\texttt{<bos>} You're a helpful Arabic assistant that answers multiple-choice questions accurately. Choose the best answer based only on the given question and options. \texttt{<start\_of\_turn>}user\\
\begin{RLtext}\footnotesize السؤال\end{RLtext}
\texttt{A.} \begin{RLtext}\footnotesize الخيار الأول\end{RLtext}\\
\texttt{B.} \begin{RLtext}\footnotesize الخيار الثاني\end{RLtext}\\
\texttt{C.} \begin{RLtext}\footnotesize الخيار الثالث\end{RLtext}\\
\texttt{D.} \begin{RLtext}\footnotesize الخيار الرابع\end{RLtext}\\[0.5em]
\texttt{<end\_of\_turn>}\\
\texttt{<start\_of\_turn>}model\\
Answer \texttt{<end\_of\_turn>}
\end{quote}
\caption{Example of a formatted prompt used for Arabic MCQ fine-tuning.}
\label{fig:formatted_prompt_structure}
\end{figure}

\begin{lstlisting}[style=pythonstyle,caption={Prompt for evaluating, refining, and filtering Arabic QA pairs.},label={lst:prompt_for_filtering_palm}]
system_prompt = """
You are an advanced NLP annotation assistant specializing in evaluating Arabic questions and answers. Your role is to classify questions, assess answers, and refine them for conciseness and accuracy.

Follow the structured guidelines for classification:
- **Step 1: Evaluate and refine the answer**, ensuring it is concise and factually correct.
- **Step 2: Determine if the question-answer pair is relevant to the Arabic culture.
    
### **Annotation Task**
You are an expert Arabic NLP QA annotator. Your task is to evaluate and refine a question-answer pair based on the following steps:

### **Step 1: Evaluate and Edit the Answer**
- **Answer Evaluation:**  
  - **Correct:** Fully and accurately answers the question.  
  - **Incorrect:** Does not answer the question or contains false information.  
  - **Partially Correct:** Provides some relevant information but is incomplete.  
- **Answer Refinement:**  
  - If correct or partially correct but **too long, vague, or redundant**, rewrite it to be **concise and precise**.

### **Step 2: Determine Arabic cultural relevance**
- **Yes:** The question explicitly refers to the Arabic culture.
- **No:** The question is about a different culture than Arabic.
- **Unsure:** It is difficult to determine whether the question refers to any specific culture.                
"""

user_prompt = f"""
### **Input Data:**

Question: {data['question']}
Answer: {data['answer']}

### **Your Response in JSON format:**
{
"answer_evaluation": "Correct" or "Incorrect" or "Partially Correct",
"corrected_answer": "Provide a concise, precise answer if needed, otherwise leave empty.",
"culture_relevance": "Yes" or "No" or "Unsure"
}
"""
\end{lstlisting}

\begin{lstlisting}[style=pythonstyle,caption={Prompt for generating 3 plausible distractors.},label={lst:prompt_for_mcq_palm}]
system_prompt = """
You are an expert in educational content creation specializing in Arabic language and culture. Your task is to convert culturally relevant question-answer pairs into multiple-choice questions (MCQs) by generating three plausible, culturally relevant, and contextually appropriate incorrect answer options (distractors) in Arabic for each question.


Requirements:
- All options must be in Arabic.
- Distractors must be plausible and relevant to the question.
- Avoid answers that are obviously incorrect, unrelated, or closely paraphrase the correct answer.
- Output only the 3 incorrect answers in the following JSON format:

JSON Output format:
{{
"A.": "",
"B": "",
"C": ""
}}
"""

user_prompt = f"""
Given the following question and its correct answer, generate 3 plausible but incorrect answer options in Arabic. 

Question: "{data['question']}"
Correct Answer: "{data['answer']}"
"""
\end{lstlisting}
\vspace{0.7cm}

\begin{lstlisting}[style=pythonstyle,caption={Prompt for identifying country.},label={lst:prompt_for_country}]

system_prompt = "You are an AI assistant for country identification."

user_prompt = """
You are an expert in Arab culture and geography. 
Given a question in Arabic, your task is to identify the most relevant Arab 
country that the question is likely referring to, either explicitly or implicitly.

Always return the name of a single Arab country in English 
(e.g., Qatar, Egypt, Saudi Arabia, UAE, Morocco, etc.).

Even if the country is not directly named, use cultural, linguistic, 
environmental, or historical clues to infer the closest matching Arab country.

Return your response in JSON format with a single field "country" 
containing only the country name.

QUESTION: "{question}"
"""
\end{lstlisting}

\begin{table}[h]
\centering
\small
\setlength{\tabcolsep}{4pt}
\renewcommand{\arraystretch}{1.1}
\begin{tabular}{ccccccc}
\toprule
\textbf{Epochs} & \textbf{LR} & \textbf{r} & \textbf{Dropout} & \textbf{Alpha} & \textbf{PalmX Dev (\%)} \\
\midrule
4 & 5e-5   & 64 & 0.15 & 16 & 79.6 \\
5 & 1.2e-4 & 64 & 0.05 & 16 & 79.8 \\
3 & 5e-5   & 64 & 0.05 & 16 & 79.8 \\
4 & 5e-5   & 64 & 0.10 & 16 & 80.2 \\
5 & 1e-4   & 32 & 0.05 & 32 & 80.4 \\
3 & 2e-4   & 64 & 0.05 & 16 & \textbf{80.5} \\
\bottomrule
\end{tabular}
\caption{PalmX Dev results from hyperparameter tuning.}
\label{tab:hyperparam-results}
\end{table}

\section{Hyperparameters}
\label{appendix:tuning_experiments}
Hyperparameter tuning varied the number of epochs (3–5), the learning rates ($5\times10^{-5}$ to $2\times10^{-4}$), the dropout rates (0.05, 0.1, 0.15), and the LoRA-specific parameters such as the rank ($r=32$ or $64$) and the scaling factor ($\alpha=16$ or $32$). Starting from a baseline, we tested higher epochs, lower learning rates, and increased dropout for regularization effects, as well as a reduced-rank, higher-$\alpha$ variant ($r=32$, $\alpha=32$). Each configuration was trained and evaluated on the PalmX Dev set to ensure consistency in reporting.

\section{Results on the Hyperparameter Tuning}
\label{appendix:tuning_results}
Fine-tuning experiments with Fanar-7B are summarized in Table~\ref{tab:hyperparam-results}. The top setup used 3 epochs, a $2\times10^{-4}$ learning rate, LoRA rank 64, dropout 0.05, and $\alpha=16$, yielding an average accuracy of 80.5 on PalmX Dev. We also observed a slight improvement when increasing the dropout to 0.1 in an earlier run with a similar configuration, and therefore incorporated this change into the top-performing setup to form our final configuration.

% \section{Submission Model Training Configuration}
% \label{appendix:submission_config}

% The full configuration used to train the submission model is shown below:

% \begin{verbatim}
% {
%     "model_name": "QCRI/Fanar-1-9B-Instruct",
%     "dataset": "palmx(train_dev)+palm+palmx-ext.",
%     "max_seq_length": 512,
%     "batch_size": 4,
%     "grad_accumulation": 4,
%     "num_epochs": 3,
%     "learning_rate": 0.0002,
%     "lora_r": 64,
%     "lora_alpha": 16,
%     "lora_dropout": 0.1,
%     "target_modules": [
%         "q_proj",
%         "v_proj",
%         "k_proj",
%         "o_proj",
%         "gate_proj",
%         "up_proj",
%         "down_proj"
%     ]
% }
% \end{verbatim}

\section{Error Analysis: Effect of Augmentation}
\label{sec:aug-analysis}

To better understand the impact of augmentation, we analyzed the subset of questions from the PalmX 2025 development set that the base model (Fanar-9B-Instruct) failed to answer correctly. Out of 500 questions, Fanar produced 138 errors.  

Finetuning on PalmX alone corrected 38 of these errors. When augmented data was included, the model solved an additional 53 questions, while losing accuracy on only 3 of the 38 cases previously resolved. In total, the augmented model recovered 88 of the 138 initially incorrect items.  

\begin{figure}[h]
    \centering
    \includegraphics[width=1\linewidth]{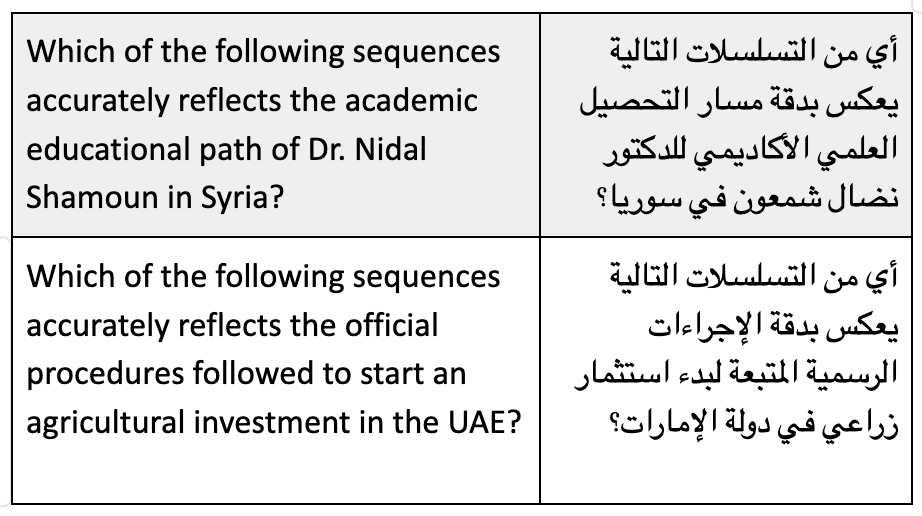}
    \caption{Questions solved by both PalmX-only and Augmentation.}
    \label{fig:palmx_and_aug}
\end{figure}

\begin{figure}
    \centering
    \includegraphics[width=1\linewidth]{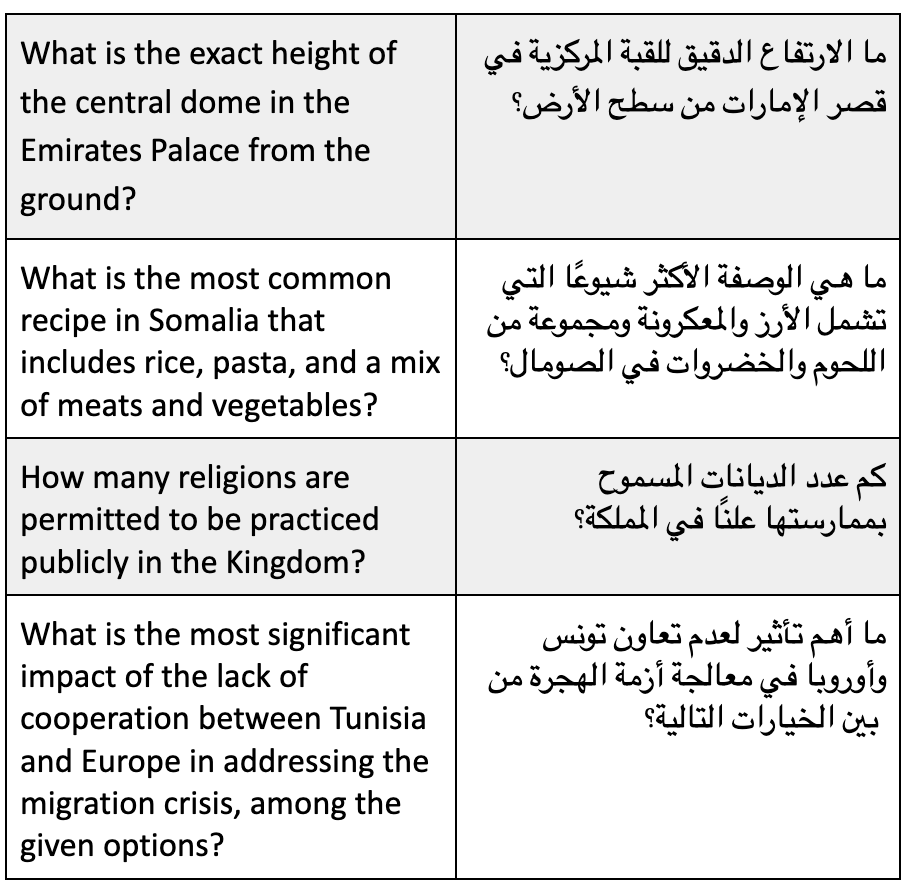}
    \caption{Questions solved only with Augmentation.}
    \label{fig:aug_only}
\end{figure}

Representative examples of these improvements are shown in Figures~\ref{fig:palmx_and_aug} and~\ref{fig:aug_only}. These illustrate how augmentation introduced broader topical coverage, especially on less-documented cultural and regional details. Without augmentation, the model remained limited to narrower knowledge encoded in PalmX.  

\section{Error Analysis: Dev vs. Test Performance}
\label{sec:dev-test-analysis}

We also examined the discrepancy between the dev and test set performance. While our model showed strong results on dev, its accuracy dropped considerably on test. To better understand this, we compared representative samples of questions from train, dev, and test.

\begin{figure}
\vspace{-0.25cm}
\centering
\includegraphics[width=1\linewidth]{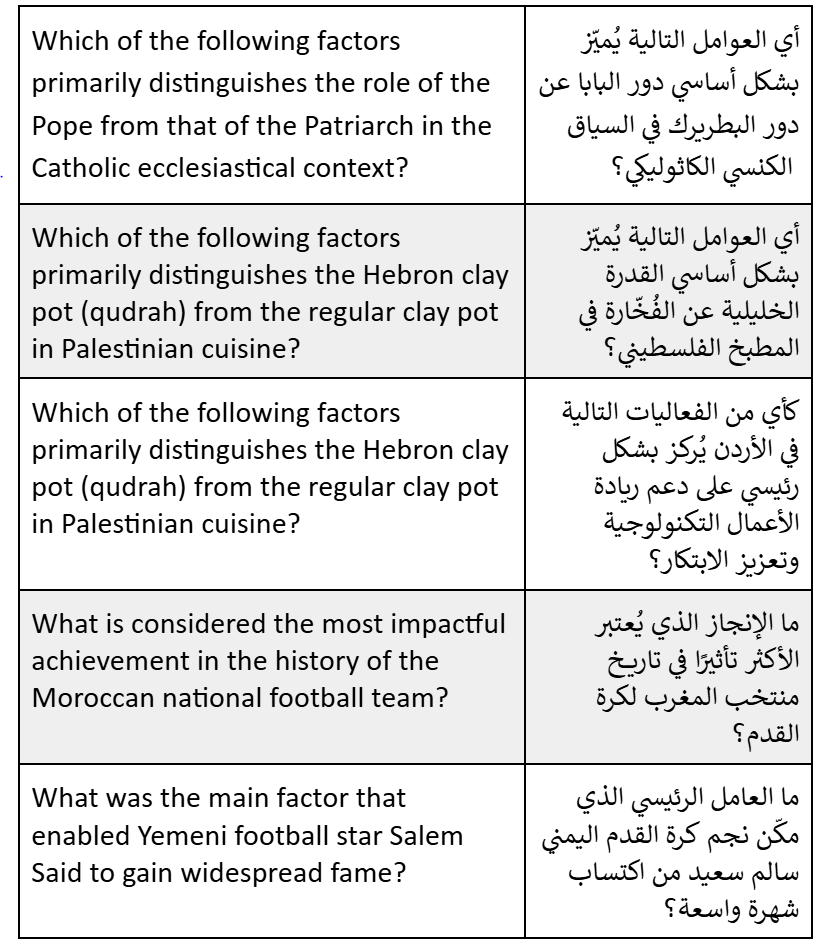}
\caption{Examples from PalmX Cultural Train Set.}
\label{fig:train_set}
\end{figure}

\begin{figure}
\centering
\includegraphics[width=1\linewidth]{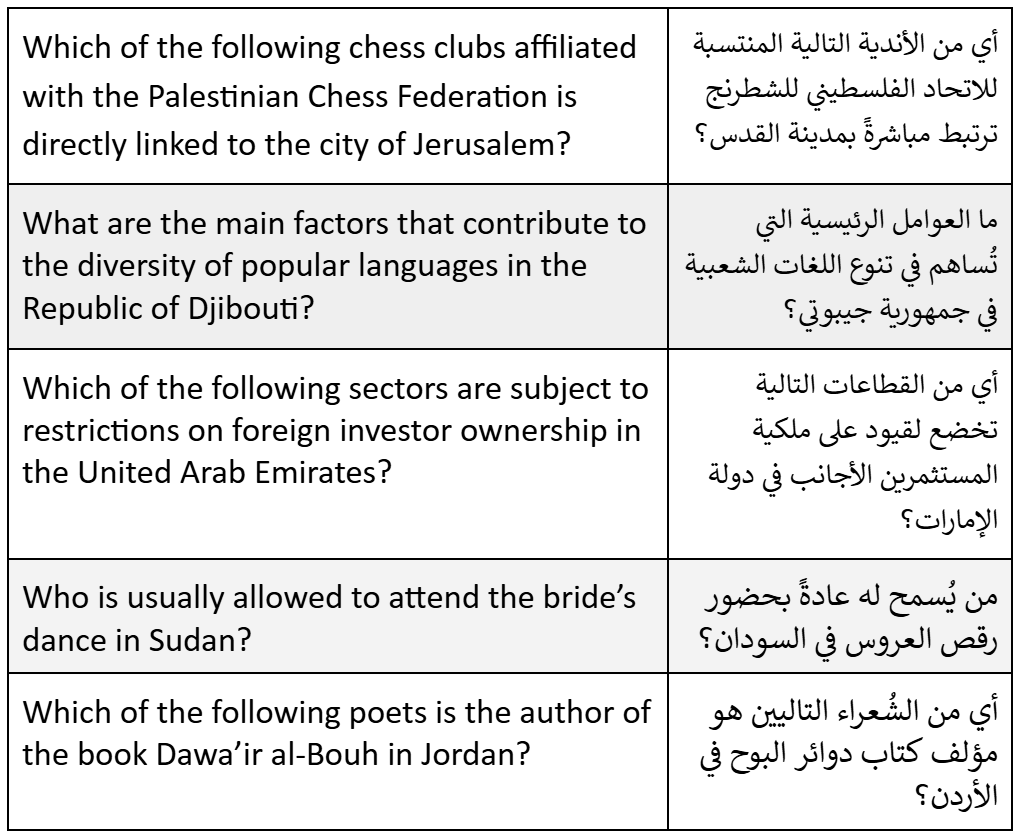}
\caption{Examples from PalmX Cultural Dev Set.}
\label{fig:dev_set}
\end{figure}

The train and dev sets are closely aligned, focusing on contemporary cultural, institutional, and social knowledge (see Figures~\ref{fig:train_set} and~\ref{fig:dev_set}). This alignment explains the stronger performance on dev: the model is effectively evaluated on material resembling what it was trained on.

\begin{figure}
\centering
\includegraphics[width=1\linewidth]{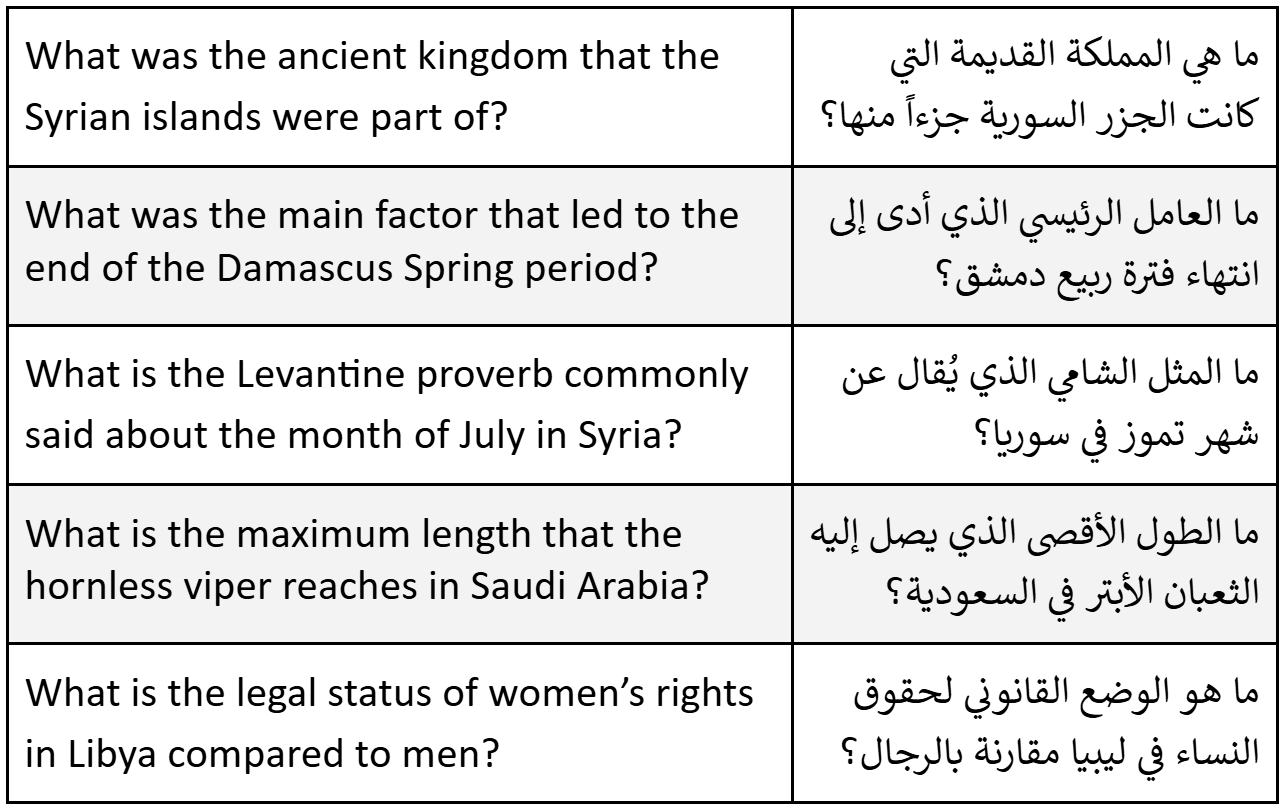}
\caption{Examples from PalmX Cultural Test Set.}
\label{fig:test_set}
\end{figure}

By contrast, the test set introduces broader and less-represented domains, including ancient history, proverbs, zoology, and legal systems (Figure~\ref{fig:test_set}). These require background knowledge beyond the distribution covered in training, explaining the observed performance drop.

It should also be noted that model development and checkpoint selection relied on dev, while the test set remained hidden, reinforcing the discrepancy.

\end{document}